\pdfoutput=1

\documentclass[11pt]{article}

\usepackage[]{EMNLP2022}

\usepackage{times}
\usepackage{latexsym}

\usepackage[T1]{fontenc}

\usepackage[utf8]{inputenc}

\usepackage{microtype}

\usepackage{inconsolata}

\usepackage{color,soul}
\usepackage{amsmath}
\DeclareMathOperator*{\argmax}{argmax}
\newcommand{\interalia}[1]{\citep[\emph{inter alia}]{#1}}

\usepackage[inline]{enumitem}

\usepackage{booktabs}
\usepackage{multirow}

\usepackage{arydshln}

\usepackage{pgfplots}
\pgfplotsset{compat=1.15}
\usepgfplotslibrary{groupplots}

\usepackage{algorithm} 
\usepackage[noend]{algorithmic}

\title{Improving Scheduled Sampling with Elastic Weight Consolidation for Neural Machine Translation}

\author{Michalis Korakakis \\
  University of Cambridge  \\
  \texttt{mk2008@cam.ac.uk} \\\And
  Andreas Vlachos \\
  University of Cambridge \\
  \texttt{av308@cam.ac.uk} \\}

\begin{document}

\maketitle
\begin{abstract}
Despite strong performance in many sequence-to-sequence tasks, autoregressive models trained with maximum likelihood estimation suffer from exposure bias, i.e.\ the discrepancy between the ground-truth prefixes used during training and the model-generated prefixes used at inference time. Scheduled sampling is a simple and empirically successful approach which addresses this issue by incorporating model-generated prefixes into training. However, it has been argued that it is an inconsistent training objective leading to models ignoring the prefixes altogether. In this paper, we conduct systematic experiments and find that scheduled sampling, while it ameliorates exposure bias by increasing model reliance on the input sequence, worsens performance when the prefix at inference time is correct, a form of catastrophic forgetting. We propose to use Elastic Weight Consolidation to better balance mitigating exposure bias with retaining performance. Experiments on four IWSLT'14 and WMT'14 translation datasets demonstrate that our approach alleviates catastrophic forgetting and significantly outperforms maximum likelihood estimation and scheduled sampling baselines.
\end{abstract}

\section{Introduction}
Autoregressive models trained with maximum likelihood estimation~(MLE) constitute the dominant approach in several sequence-to-sequence tasks, such as machine translation~\citep{DBLP:journals/corr/BahdanauCB14}, text summarization~\citep{see-etal-2017-get}, and conversational modeling~\citep{DBLP:journals/corr/VinyalsL15}. However, this paradigm suffers from exposure bias~\citep{DBLP:conf/nips/BengioVJS15, DBLP:journals/corr/RanzatoCAZ15}, i.e.\ during training the model generates tokens by conditioning on the ground-truth prefixes, while at inference time model-generated prefixes are used instead. Since the model is never exposed to its own errors, if a token is mistakenly generated during inference the error will be propagated along the sequence~\citep{DBLP:journals/jmlr/RossGB11}. Prior work has attributed to exposure bias various forms of text degeneration, such as repetitiveness, incoherence, and hallucinations~\citep{DBLP:conf/iclr/HoltzmanBDFC20, wang-sennrich-2020-exposure}, i.e.\ outputs which contain information irrelevant to the input sequence.

\citet{DBLP:conf/nips/BengioVJS15} introduced scheduled sampling to address exposure bias in MLE-trained autoregressive models. Scheduled sampling uses a stochastic mixture of ground-truth and model-generated prefixes during training, thereby allowing the model to learn how to recover from its own errors. While various alternatives to MLE have been proposed to mitigate exposure bias~\interalia{DBLP:journals/corr/RanzatoCAZ15, wiseman-rush-2016-sequence, shen-etal-2016-minimum, DBLP:conf/iclr/BahdanauBXGLPCB17}, scheduled sampling remains one of the most popular due to its simplicity and performance improvements in many conditional sequence generation tasks~\citep{DBLP:conf/nips/BengioVJS15, du-ji-2019-empirical, zhang-etal-2019-bridging, DBLP:conf/icml/LiL21a}.

Conversely, other studies have reported that using scheduled sampling may hurt performance~\citep{DBLP:conf/iclr/LeblondAOL18, mihaylova-martins-2019-scheduled}. The dominant hypothesis for these negative results is that scheduled sampling creates models that are more likely to recover from their own mistakes by training them to ignore the prefixes entirely~\citep{DBLP:journals/corr/Huszar15}. However, no attempt has been made to empirically assess this hypothesis. Thus it is still unclear how scheduled sampling affects training.

In this paper, we provide insights into the working mechanisms of scheduled sampling. Following~\citet{DBLP:journals/corr/abs-2010-10907}, we apply Layerwise Relevance Propagation~(LRP)~\citep{10.1371/journal.pone.0130140} to quantify the contributions of the input sequence and the prefix during inference, and empirically evaluate the hypothesis of~\citet{DBLP:journals/corr/Huszar15}. We find that models trained with scheduled sampling increase their reliance on the input sequence, and therefore mitigate exposure bias by depending less on the potentially incorrect model-generated prefix. However, we also observe that this has the side-effect of worsening the predictions when the model-generated prefix is correct. To address this form of catastrophic forgetting~\citep{FRENCH1999128}, we propose using Elastic Weight Consolidation~(EWC)~\citep{DBLP:journals/corr/KirkpatrickPRVD16} to regularize the parameter updates in scheduled sampling so that the performance is not affected when the prefix is correct.

Experiments on the commonly-used IWSLT'14 German-English, IWSLT'14 Vietnamese-English, WMT'14 English-German, and WMT'14 English-French translation datasets show that our proposed method mitigates catastrophic forgetting and significantly improves translation performance, in terms of BLEU~\citep{papineni-etal-2002-bleu}, over MLE, standard scheduled sampling, and a recently proposed scheduled sampling variant~\citep{zhang-etal-2019-bridging}. Importantly, performance gains occur in both long short-term memory~(LSTM)~\citep{DBLP:journals/neco/HochreiterS97} and Transformer~\citep{DBLP:conf/nips/VaswaniSPUJGKP17} models, and across different annealing schedules, demonstrating that our proposed EWC-regularized scheduled sampling is more robust and easier to tune. Finally, human evaluation showcases that our method can improve both translation adequacy and fluency.

\section{Scheduled Sampling}
Autoregressive models estimate the conditional probability of the output $\boldsymbol{y}$ given the input $\boldsymbol{x}$ one token at a time in a monotonic fashion:

\begin{align}
\label{eq:prob}
P(\boldsymbol{y} \mid \boldsymbol{x}) = \prod_{t=1}^{T}p(y_t  \mid \boldsymbol{y}_{<t},\boldsymbol{x};\theta),
\end{align}
\noindent where $y_{t}$ is the t-th token in $\boldsymbol{y}$, $\boldsymbol{y}_{<t}$ denotes all previous tokens, and $\theta$ is the set of model parameters.

Given a dataset $\mathcal{D} = \{  (\boldsymbol{x}^{(i)}, \boldsymbol{y}^{(i)}) \}_{i=1}^{N}$ of input-output pairs, the standard approach to optimize the parameters $\theta$ of an autoregressive model entails maximizing the conditional log-likelihood:

\begin{gather}
\hat{\theta}_{\text{MLE}} = \argmax_{\theta}\mathcal{L}(\theta;\mathcal{D}), \\
\mathcal{L}(\theta; \mathcal{D}) = \sum_{i=1}^{N}\sum_{t=1}^{L^{(i)}}\log p(y_t^{(i)} \mid \boldsymbol{y}^{(i)}_{<t},\boldsymbol{x}^{(i)};\theta).
\label{eq:tf}
\end{gather}

Here $i$ indicates the i-th output sequence in the dataset and $L^{(i)}$ is the length of the i-th output sequence. This training objective is known as teacher-forcing~\citep{DBLP:journals/neco/WilliamsZ89}, since the model conditions on the ground-truth prefix $\boldsymbol{y}^{(i)}_{<t}$ to generate the token $y^{(i)}_{t}$. However, at inference time, the model generates the token $\hat y_{t}$ by conditioning on its own outputs, i.e.\ $\boldsymbol{\hat{y}}_{<t}$ instead of $\boldsymbol{y}_{<t}$, which creates a discrepancy between training and inference known as the exposure bias problem~\citep{DBLP:conf/nips/BengioVJS15, DBLP:journals/corr/RanzatoCAZ15}. 

\begin{algorithm}[!t]
  \caption{Scheduled Sampling}
  \label{alg:seq_ss}
{\fontsize{10.5}{10}\selectfont

  \begin{algorithmic}[1]
  
    \STATE {\bf Input}: Dataset $\mathcal{D}$
    \STATE Initialize $a = 1$
        \FOR {$i = 1,\ldots,I$}
    \REPEAT 

    \FOR {$t = 1,\ldots,T$}
    \STATE $\tilde{y}_t$ = $\begin{cases}\displaystyle 
                y_t & \text{with  $a$} \\ 
                \hat{y}_t \sim p_{\theta}(y_t | \tilde{y}_{1:t-1}, x) & \text{with  $1-a$} 
                \end{cases}$
    \STATE $L({\theta}) = L({\theta}) + \log p_{\theta}(y_t | \tilde{y}_{1:t-1}, x)$ \label{loss}
    \ENDFOR
    \UNTIL{$B$ \textbf{times}}

    \STATE ${\theta} = {\theta} + \eta\cdot\nabla_{{\theta}}L({\theta})$\label{params}
    \STATE $a = \textsc{Schedule}(a)$ 
    \ENDFOR
  \end{algorithmic}}
\end{algorithm}

\citet{DBLP:conf/nips/BengioVJS15} introduced scheduled sampling to mitigate the above-mentioned discrepancy between MLE training and inference. Scheduled sampling uses the same training objective as teacher-forcing~(Equation~\ref{eq:tf}), the only difference being that the conditioning prefixes $\boldsymbol{\tilde{y}}^{(i)}_{<t}$ are a stochastic mixture of ground-truth $\boldsymbol{y}^{(i)}_{<t}$ and model-generated prefixes $\boldsymbol{\hat{y}}^{(i)}_{<t}$:

\begin{align}
\mathcal{L}(\theta; \mathcal{D}) = \sum_{i=1}^{N}\sum_{t=1}^{L^{(i)}}\log p(y_t^{(i)} \mid \boldsymbol{\tilde{y}}^{(i)}_{<t},\boldsymbol{x}^{(i)};\theta).
\label{eq:ss}
\end{align}

An annealing schedule is used to gradually decrease the probability $a$ of conditioning using the ground-truth prefix during training. Typically, for each mini-batch $b$, $a$ is decreased using the following annealing schedules:

\begin{itemize}[noitemsep]
    \item \textbf{Linear}: $a = \max (a - kb, 0)$
    \item \textbf{Exponential}: $a = k^b$
    \item \textbf{Inverse sigmoid}: $a = k/(k+\exp (b/k))$
\end{itemize}
Here $k$ is a hyperparameter which controls the speed of decay of $a$ in each schedule. Algorithm~\ref{alg:seq_ss} summarizes training with scheduled sampling.

\section{Analysis of Scheduled Sampling}\label{sec:analysis-ss}
In this section, we propose two ways to investigate how scheduled sampling works. First, we use Layerwise Relevance Propagation~(LRP)~\citep{10.1371/journal.pone.0130140, DBLP:journals/corr/abs-2010-10907} to examine the contributions of the input sequence and the prefix during inference, and thus assess the hypothesis of~\citet{DBLP:journals/corr/Huszar15}. Then we treat training with scheduled sampling as a form of progressive fine-tuning, by assuming that the model-generated prefixes can be viewed as a downstream task we adapt the model to after it is trained with the ground-truth prefixes. To this end, we apply teacher-forcing at inference time to quantify the impact of catastrophic forgetting on models trained with scheduled sampling.

\subsection{Prefixes under Scheduled Sampling}\label{subsec:lrp}
\citet{DBLP:journals/corr/Huszar15} argued that scheduled sampling is an inappropriate training objective since it learns models that ignore the prefixes. This limitation arises because the model-generated outputs correspond to a distribution that is different from the ground-truth sequences the model is trained to generate. Therefore training might not converge to the correct model even as the dataset and the capacity of the model increase indefinitely.

To empirically verify this hypothesis, we apply LRP to quantify the influence of the input sequence and the prefix during prediction when the model conditions using \begin{enumerate*}[label=(\roman*)] \item random prefixes that are valid in the target language but semantically unrelated to the input sequence, and \item model-generated prefixes\end{enumerate*}. The intuition behind this experiment is that if scheduled sampling creates models that ignore the prefixes, then the input sequence and the prefix contributions should remain relatively stable across the two prefix types.

LRP quantifies the proportion of each token's influence during prediction, and thus it has been used to analyze the effect of source and target contexts~\citep{ding-etal-2017-visualizing, DBLP:journals/corr/abs-2010-10907}. In terms of explainability and reliability, it has been shown to outperform related methods in text classification~\citep{arras-etal-2017-explaining, poerner-etal-2018-evaluating, arras-etal-2019-evaluating}. Concretely, given an input sequence token $x_i$ and a prefix token $y_j$, LRP computes the relevance scores $r_t(x_i)$ and $r_t(y_j)$ at every output generation step $t$ by back-propagating from the output to the input. Importantly, the total relevance scores for each generated token are equal to 1:
\begin{align}
\sum\limits_{i}r_t(x_i)\!+\!\sum\limits_{j=1}^{t-1}r_t(y_j)\!=\!1.\label{eq:lrp_contrib}
\end{align}
In the general case, the relevance score $r_{j}^{(l)}$ for the $j$-th neuron in layer $l$ is computed as:\footnote{Since we use implementation of~\citet{DBLP:journals/corr/abs-2010-10907}, we adopt the LRP-$\alpha\beta$ rule~\citep{10.1371/journal.pone.0130140, DBLP:conf/icann/BinderMLMS16} for computing relevance scores.}
\begin{align}
r_{j}^{(l)} = \sum_{k=1}^{K} \left(\alpha \frac{z_{jk}^{+}}{\sum_{i} z_{ji}^{+}} - \beta \frac{z_{jk}^{-}}{\sum_{i} z_{ji}^{-}}\right)r_{k}^{(l+1)},
\end{align}
\noindent where $z_{jk} = x_{j}w_{jk}$. Here $K$ denotes the total number of neurons at each layer, $j$ and $k$ are neurons at two consecutive layers $l$ and $l + 1$, $x_{j}$ is the activation of neuron $j$ at layer $l$, $w_{jk}$ is the weight connecting neurons $j$ and $k$, $r(l+1)$ is the relevance score at layer $(l+1)$, $+$ and $-$ indicate positive and negative contributions to $r(l+1)$, and finally, $\alpha$ and $\beta$ are hyperpameters that determine the importance of the positive~(+) and negative~(-) contributions.

\begin{algorithm}[!t!]
\caption{Scheduled Sampling as Fine-Tuning} \label{algo:progr_fine_tune}
{\fontsize{10.5}{10}\selectfont

\begin{algorithmic}[1]
\STATE {\bf Input}: Dataset $\mathcal{D}$

\STATE {Initialize $a = 1$}
    \FOR {$i = 1, \ldots, I$} 
    \STATE {Initialize $\tilde{\mathcal{D}}, \tilde{\mathcal{D}}_{cond} = \emptyset$}
    \REPEAT 

        \STATE Sample example $(x, y)$ from $\mathcal{D}$
        \FOR {$t = 1,\dots,T$}
        \STATE $\tilde{y}_t$ = $\begin{cases}\displaystyle 
            y_t & \text{with  $a$} \\ 
            \hat{y}_t \sim p_{\theta}(y_t | \tilde{y}_{1:t-1}, x)  & \text{with  $1-a$} 
        \end{cases}$
    \ENDFOR
    \STATE Add $(x, y)$ to $\tilde{\mathcal{D}}$ and $(x, \tilde{y})$ to $\tilde{\mathcal{D}}_{cond}$
    
    \UNTIL{$B$ \textbf{times}}

    \STATE Compute loss on $\tilde{\mathcal{D}}$ conditioning with $\tilde{\mathcal{D}}_{cond}$\STATE $a\gets\textsc{Schedule}(a)$
\ENDFOR
\end{algorithmic}}
\end{algorithm}

\subsection{Scheduled Sampling as Fine-Tuning}\label{subsec:quant_cf} 
Our hypothesis is that scheduled sampling can be viewed as progressively fine-tuning a model using its own predictions as prefixes to address exposure bias~(Algorithm~\ref{algo:progr_fine_tune}). A commonly occurring issue is that during fine-tuning the model's ability to retain previously learned knowledge is negatively impacted, a phenomenon known as catastrophic forgetting~\citep{FRENCH1999128}.

To this end, inspired by~\citet{wu-etal-2018-beyond} who applied teacher-forcing during inference to investigate the influence of exposure bias in NMT, we use the ground-truth prefix $\boldsymbol{y}_{<t}$ instead of the model-generated prefix $\boldsymbol{\hat{y}}_{<t}$ to generate the output token $\boldsymbol{\hat y_{t}}$. If there is no catastrophic forgetting, then the output quality of models trained with scheduled sampling is expected to be similar to that of MLE-trained models when ground-truth tokens are used during inference.

\section{Experiments}\label{sec:experiments}
In this section we apply the methods discussed in Section~\ref{sec:analysis-ss} to examine if training with scheduled sampling leads to models ignoring the prefixes, and also determine whether scheduled sampling causes catastrophic forgetting.

\subsection{Experimental Setup}
\paragraph{Data}
Following prior work exploring scheduled sampling in NMT~\citep{goyal-etal-2017-differentiable, xu-etal-2019-differentiable}, we conduct experiments on two IWSLT'14 translation datasets, the German-English (DE-EN) and Vietnamese-English (VI-EN). We use byte pair encoding~(BPE)~\citep{sennrich-etal-2016-neural} to encode sentences. We report cased, tokenized BLEU~\citep{papineni-etal-2002-bleu}.\footnote{\url{https://github.com/moses-smt/mosesdecoder/blob/master/scripts/generic/multi-bleu.perl}}

\paragraph{Models}
We use LSTM~\citep{DBLP:journals/neco/HochreiterS97} and Transformer~\citep{DBLP:conf/nips/VaswaniSPUJGKP17} models trained with default architectures and hyperparameters from fairseq~\citep{ott-etal-2019-fairseq}. Models are first trained with MLE and then fine-tuned using scheduled sampling. For scheduled sampling, we use the inverse sigmoid annealing schedule for all experiments unless otherwise stated, since we found it to perform better than the linear and exponential schedules. We set all hyperparameters according to the performance on the validation set of each task. We report the mean and standard deviation over 5 runs with different random seeds.

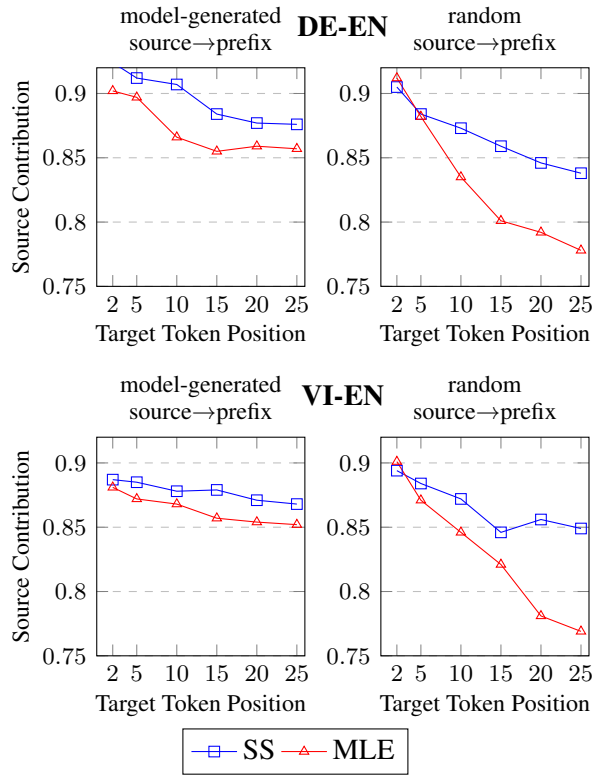
\begin{figure}[!t]
  \begin{center}
    \begin{tikzpicture}
    \begin{groupplot}[group style={
            group size=2 by 2,
            ylabels at=edge left,
            vertical sep= 2cm
        }, 
        height=0.28\textwidth,
        width=0.27\textwidth,
        xmin=0, xmax=26,
        ymin=0.75, ymax=0.92,
        xtick={2, 5, 10, 15, 20, 25},
        ymajorgrids=true,
        xlabel={Target Token Position},
        ylabel style={align=center},
        ylabel={Source Contribution},
        ylabel shift=-0.5ex,
        xlabel shift=-0.5ex,
        grid style=dashed,
        label style={font=\small},
        title style={yshift=-1ex},
        tick label style={font=\small},
        legend to name=named,
        legend columns=-1
    ]
    \nextgroupplot[align=center, title={model-generated \\ 
    source\textrightarrow prefix}, style={font=\small}]
     \addplot[color=blue, mark=square] coordinates {(2,0.924) (5,0.912) (10,0.907) (15,0.884) (20,0.877) (25,0.876)};
    \addplot[color=red, mark=triangle]  coordinates {(2,0.902) (5,0.897) (10,0.866) (15,0.855) (20,0.859) (25,0.857)};
    \nextgroupplot[align=center, title={random \\ 
    source\textrightarrow prefix}, style={font=\small}]
    \addplot[color=blue, mark=square] coordinates {(2,0.905) (5,0.884) (10,0.873) (15,0.859) (20,0.846) (25,0.838)};
    \addplot[color=red, mark=triangle] coordinates {(2,0.912) (5,0.882) (10,0.835) (15,0.801) (20,0.792) (25,0.778)};
    \nextgroupplot[align=center, title={model-generated \\ 
    source\textrightarrow prefix}, style={font=\small}]        \addplot[color=blue, mark=square] coordinates {(2,0.887) (5,0.885) (10,0.878) (15,0.879) (20,0.871) (25,0.868)};
    \addplot[color=red, mark=triangle]  coordinates {(2,0.881) (5,0.872) (10,0.868) (15,0.857) (20,0.854) (25,0.852)};
    \nextgroupplot[align=center, title={random \\ 
    source\textrightarrow prefix}, style={font=\small}]
    \addplot[color=blue, mark=square] coordinates {(2,0.894) (5,0.884) (10,0.872) (15,0.846) (20,0.856) (25,0.849)};
    \addplot[color=red, mark=triangle] coordinates {(2,0.901) (5,0.871) (10,0.846) (15,0.821) (20,0.781) (25,0.769)};
    \addlegendimage{blue, mark=square}
    \addlegendentry[color=black]{SS}
    \addlegendentry[color=black]{MLE}
    \end{groupplot}
    \node (title) at ($(group c1r1.center)!0.5!(group c2r1.center)+(0,2cm)$) {\textbf{DE-EN}};
    \node (title) at ($(group c1r2.center)!0.5!(group c2r2.center)+(0,2cm)$) {\textbf{VI-EN}};
    \end{tikzpicture}
    \ref{named}
    \caption{Source contributions with model-generated and random prefixes.}\label{fig:ferenc_analysis_lrp}
  \end{center}
\end{figure}

\subsection{Results}\label{subsec:exp_results}
\paragraph{Prefixes under Scheduled Sampling}
Figure~\ref{fig:ferenc_analysis_lrp} shows the source contributions for the random and model-generated prefixes. Following the experimental setup of~\citet{DBLP:journals/corr/abs-2010-10907}, we randomly choose 100 source-target pairs from each test set with the same length in the source and target and average all results, as this renders the analyses of the contributions more reliable. Furthermore, we use the Transformer model to quantify contributions since the implementation of LRP by~\citet{DBLP:journals/corr/abs-2010-10907} is tailored to this particular neural architecture. Figure~\ref{fig:ferenc_analysis_lrp} illustrates that at the beginning, MLE-trained models tend to use the source more, however, as the target prefix becomes longer, they rely less on the source and more on the prefix (especially for the random prefixes), given that:

\begin{align}
\sum\limits_{j=1}^{t-1}r_t(y_j) = 1 - \sum\limits_{i}r_t(x_i).
\label{eq:lrp_eq}
\end{align}

\noindent This behavior is in agreement with prior work that has attributed the increasing reliance on the prefix to teacher-forcing~\citep{wang-sennrich-2020-exposure, DBLP:journals/corr/abs-2010-10907}. Conversely, scheduled sampling, which mitigates exposure bias, promotes the usage of source information to generate a target token, as the decrease in the influence of source contributions is less drastic compared to MLE-trained models across the two prefix types. This suggests that scheduled sampling results in models that tend to ignore the prefix more than MLE-trained models, thus confirming the hypothesis of~\citet{DBLP:journals/corr/Huszar15}. However, in contrast to~\citet{DBLP:journals/corr/Huszar15}, we argue that this is a positive outcome in the context of NMT, i.e.\ the model-generated prefix is often incorrect, but the source that is being translated does not change. Therefore, learning to rely on it more is a reasonable approach to alleviate exposure bias.

\begin{table}[!t]
\setlength{\tabcolsep}{4pt}
\centering
\small
\begin{tabular}{lccccccc}
\toprule
& \multicolumn{3}{c}{\textbf{DE-EN}} & & \multicolumn{3}{c}{\textbf{VI-EN}}   \\
\cmidrule{2-4}  \cmidrule{6-8} 
\textbf{Method} & MP & TF & {${\Delta}$} & & MP & TF & ${\Delta}$\\
\midrule
\textbf{LSTM} \\
\quad {MLE} & 26.72 & \textbf{28.67} & +1.95  & & 22.64 & \textbf{24.98} & +2.34\\
\quad {SS} & \textbf{27.35}  & 23.59 &  -3.76  & & \textbf{23.58} & 18.92 & -4.66\\
\midrule
\multicolumn{2}{l}{\textbf{Transformer}}\\
\quad {MLE} & 34.08 & \textbf{35.96} & +1.88  & & 27.13 & \textbf{29.14} & +2.01\\
\quad {SS}& \textbf{34.65}  & 29.84 & -4.81  & & \textbf{27.57} & 22.42 & -5.15 \\
\bottomrule
\end{tabular}
\caption{BLEU scores by conducting inference with two approaches, namely, \begin{enumerate*}[label=(\roman*)]
\item by conditioning using model-generated prefixes~(MP), and \item with teacher-forcing~(TF), i.e.\ feeding the ground-truth as an input to generate the output token.\end{enumerate*}}\label{tab:tf-inference}
\end{table}

\paragraph{Scheduled Sampling as Fine-Tuning}
Table~\ref{tab:tf-inference} empirically confirms the occurrence of catastrophic forgetting in models trained with scheduled sampling. We observe that when models condition using the ground-truth prefixes, the quality of generations drops heavily in terms of BLEU across both languages and neural architectures, suggesting that scheduled sampling results in forgetting how to predict when the prefix is correct. In particular, on DE-EN the BLEU score reduction is 3.76 for the LSTM model and 4.81 for the Transformer. Similarly, on VI-EN the BLEU score degrades by 4.66 and by 5.15. On the other hand, we see that in MLE-trained models, using teacher-forcing at inference time leads to consistent BLEU score improvements over conditioning with model-generated prefixes. Specifically, on DE-EN the BLEU score gain is 1.95 for the LSTM model and 1.88 for the Transformer, and on VI-EN the BLEU score increases by 2.34 and by 2.01.

\section{Elastic Weight Consolidation for Scheduled Sampling}
Section~\ref{sec:experiments} shows that even though scheduled sampling addresses exposure bias by increasing model reliance on the input sequence, it also leads to output degradation due to catastrophic forgetting~\citep{FRENCH1999128} when the model-generated prefix is correct at inference time. Elastic Weight Consolidation~(EWC)~\citep{DBLP:journals/corr/KirkpatrickPRVD16} is an effective regularization-based approach for mitigating catastrophic forgetting during fine-tuning with successful applications in NMT~\citep{thompson-etal-2019-overcoming, saunders-etal-2019-domain}.

To this end, we use EWC to regularize conditioning with the model-generated prefixes ${\hat{y}_{<t}}$ to avoid forgetting using the ground-truth ${y_{<t}}_{j}$ as prefixes. EWC penalizes changes to the parameters which are important for the ground-truth prefixes, while adapting more the parameters that are less important:
\begin{multline}
\mathcal{L}_{\text{EWC}}(\theta^{\hat{y}_{<t}}; \mathcal{D}^{\hat{y}_{<t}}) = \mathcal{L}(\theta^{\hat{y}_{<t}}; \mathcal{D}^{\hat{y}_{<t}})\\ 
+\lambda \sum_j F_j (\theta^{\hat{y}_{<t}}_j - \theta^{y_{<t}}_{j})^2,
\label{eq:ewc}
\end{multline}
\noindent where $\mathcal{L}(\theta^{\hat{y}_{<t}}; \mathcal{D}^{\hat{y}_{<t}})$ is the loss~(Equation~\ref{eq:ss}) over the model-generated prefixes, $\lambda$ is a hyperparameter which determines the importance of conditioning with ${y_{<t}}$, $\theta^{y_{<t}}_j$ are the parameters of the initial model trained with ${y_{<t}}$, $\theta^{\hat{y}_{<t}}_j$ are the parameters of the model trained with $\hat{y}_{<t}$, and $F_j=E \big[ \nabla^2 L(\theta^{y_{<t}}_{j})\big]$ is an estimate of how important the parameters $\theta^{y_{<t}}_j$ are to ${y_{<t}}$. This can be approximated via the empirical Fisher information matrix ~\citep{DBLP:journals/corr/Martens14}, which requires samples from ${y_{<t}}$. Concretely, using the EWC regularizer, we modify the Lines~\ref{loss} and~\ref{params} of Algorithm~\ref{alg:seq_ss} to penalize changes in model parameters associated with conditioning using the ground-truth.

\section{Experiments with EWC}
In this section, we perform experiments with the proposed EWC-regularized scheduled sampling variant following the setup described in Section~\ref{sec:experiments}. To further demonstrate the effectiveness of our method, we also conduct experiments on two large-scale WMT'14 translation datasets, the English-German~(EN-DE) and English-French~(EN-FR). In addition, we report results using the sentence-oracle scheduled sampling variant proposed by~\citet{zhang-etal-2019-bridging}, which conditions on model-generated prefixes obtained via beam search.

\begin{table*}
\centering
\begin{tabular}{lccccc}
	\toprule
			& \multicolumn{2}{c}{\textbf{IWSLT'14}} & \multicolumn{2}{c}{\textbf{WMT'14}} \\\cmidrule(lr){2-3} \cmidrule(l){4-6} 
			\textbf{Method}   & \multicolumn{1}{c}{DE-EN} & \multicolumn{1}{c}{VI-EN} & \multicolumn{1}{c}{EN-DE} & \multicolumn{1}{c}{EN-FR} \\ \midrule
		\textbf{LSTM} \\
			\quad {MLE} & {26.72{$\pm$0.21}} & {22.64{$\pm$0.16}} & 25.43{$\pm$0.28} & 32.43{$\pm$0.33}  \\ 
			
			\quad {SS}  & {27.35}{$\pm$0.32} & {23.58}{$\pm$0.33} & 26.08{$\pm$0.22} & 32.76{$\pm$0.26}  \\
		    \quad\quad {+ EWC} & {{27.97}}{$\pm$0.24} & {{24.31}}{$\pm$0.27} & {26.68}{$\pm$0.25} & {33.18}{$\pm$0.31} \\
            \quad {Sent.} & 27.68{$\pm$0.23} & 23.91{$\pm$0.25} & 26.52{$\pm$0.19} & 33.14{$\pm$0.25}\\ 
            \quad\quad {+ EWC} & {28.14}{$\pm$0.19} & {24.33}{$\pm$0.17} & {26.83}{$\pm$0.27} & {33.48}{$\pm$0.29} \\
            
            \midrule
            
            \multicolumn{2}{l}{\textbf{Transformer}}\\
            \quad {MLE}  & {34.08{$\pm$0.15}} & {27.13{$\pm$0.18}} & 27.26{$\pm$0.16} & 38.92{$\pm$0.19} \\ 
            
            \quad {SS} & {34.65}{$\pm$0.19} & {27.57}{$\pm$0.26} & 27.92{$\pm$0.11} & 39.86{$\pm$0.24} \\
            \quad\quad {+ EWC} & {35.19}{$\pm$0.21}  & {28.16}{$\pm$0.26} & {28.44}{$\pm$0.15} & {40.33}{$\pm$0.17} \\
            \quad {Sent.} & 34.89{$\pm$0.15}  & 27.82{$\pm$0.17} & 28.34{$\pm$0.16} & 40.51{$\pm$0.20}  \\
            \quad\quad {+ EWC} & {35.24}{$\pm$0.13} & {28.11}{$\pm$0.16} & {28.61}{$\pm$0.10} & {40.83}{$\pm$0.25} \\
    \bottomrule
\end{tabular}
\caption{Average BLEU scores and standard deviation on test sets for models trained on DE-EN,VI-EN, EN-DE, and EN-FR.}
\label{tab:ewc+ss_results}
\end{table*}

\begin{table}[t]
\centering
\begin{tabular}{lcc}
\toprule
\textbf{Model}    & \textbf{Adequacy} & \textbf{Fluency} \\  \hline
MLE                 & 4.23     & 4.69        \\ 
SS                  & 4.35     & 4.78      \\
\quad {+ EWC}            & \textbf{4.41}     & \textbf{4.82}        \\
\bottomrule
\end{tabular}
\caption{Human evaluation on adequacy and fluency for DE-EN.}
\label{tab:human_eval}
\end{table}

\paragraph{Automatic Evaluation}
Table~\ref{tab:ewc+ss_results} shows that by using the EWC-regularized scheduled sampling variant we obtain better BLEU scores across all translation datasets and neural architectures. Specifically, for the LSTM model, our method outperforms standard scheduled sampling by 0.62, 0.73, 0.60, and 0.42 BLEU points on DE-EN, VI-EN, EN-DE, and EN-FR, respectively. Similarly, for the Transformer model, it outperforms standard scheduled sampling by 0.54, 0.59, 0.52, and 0.47 BLEU points. Furthermore, we also observe that our method improves over the sentence oracle scheduled sampling variant. For the LSTM model, it outperforms sentence oracle by 0.46, 0.42, 0.31, and 0.34 BLEU points, and for the Transformer model, by 0.35, 0.29, 0.27 and 0.32 BLEU points. Improvements are significant with $p$ < 0.01.\footnote{We conduct statistical significance testing following~\citet{collins-etal-2005-clause}.}

\paragraph{Human Evaluation} 
To further examine the differences between MLE, scheduled sampling, and our proposed EWC-regularized variant, we conduct a human evaluation. We ask annotators to rate from 1 to 5 in terms of fluency and adequacy 100 randomly sampled DE-EN translations generated by the Transformer model. Table~\ref{tab:human_eval} shows the results of the human evaluation. Compared with the MLE baseline, both scheduled sampling and our EWC-regularized variant improve adequacy by 0.12 and 0.18, respectively. For fluency, scheduled sampling achieves a 0.09 score improvement, and our variant a 0.13 score increase.

\begin{table}[!t]
\setlength{\tabcolsep}{3pt}
\centering
\small
\begin{tabular}{lccccccc}
\toprule
& \multicolumn{3}{c}{\textbf{DE-EN}} & & \multicolumn{3}{c}{\textbf{VI-EN}}   \\
\cmidrule{2-4}  \cmidrule{6-8} 
\textbf{Method} & MP & TF & {${\Delta}$} & & MP & TF & ${\Delta}$\\
\midrule
\textbf{LSTM} \\
\quad {MLE} & 26.72 & \textbf{28.67} & +1.95  & & 22.64 & \textbf{24.98} & +2.34\\
\quad {SS} & {27.35}  & 23.59 &  -3.76  & & {23.58} & 18.92 & -4.66\\ \hdashline
\quad\quad {+ EWC} & \textbf{{27.97}}  & 26.90 & -1.07  & & \textbf{24.31} & 22.93 & -1.38 \\

\midrule
\multicolumn{2}{l}{\textbf{Transformer}}\\
\quad {MLE} & 34.08 & \textbf{35.96} & +1.88  & & 27.13 & \textbf{29.14} & +2.01\\
\quad {SS}& {34.65}  & 29.84 & -4.81  & & {27.57} & 22.42 & -5.15 \\ \hdashline
\quad\quad {+ EWC}& \textbf{{35.19}}  & 33.17 & -2.02  & & \textbf{28.16} & 26.05 & -2.11 \\

\bottomrule
\end{tabular}
\caption{BLEU scores by conducting inference with two approaches, namely, \begin{enumerate*}[label=(\roman*)]
\item by conditioning using model-generated prefixes~(MP), and \item with teacher-forcing~(TF)\end{enumerate*}. Results for MLE and SS repeat numbers from Table~\ref{tab:tf-inference}.}\label{tab:tf-inference_with_ewc}
\end{table}

\paragraph{Catastrophic Forgetting} We investigate whether EWC addresses the catastrophic forgetting problem in models trained with scheduled sampling by repeating the teacher-forcing experiment discussed in Section~\ref{sec:analysis-ss}. Table~\ref{tab:tf-inference_with_ewc} shows less catastrophic forgetting occurring compared to the results obtained with standard scheduled sampling. In particular, on DE-EN the reduction in BLEU for standard scheduled sampling is 3.76 and 4.81, whereas with EWC the reduction is 1.07 and 2.02. Similarly, on VI-EN for standard scheduled sampling the BLEU score decreases by 4.66 and by 5.15, while for scheduled sampling with EWC by 1.38 and 2.11. This suggests that using EWC explicitly addresses the catastrophic forgetting phenomenon.

\begin{figure*}[!t]
  \begin{center}
\begin{tikzpicture}
  \begin{groupplot}[
      group style = {
        group size = 3 by 2,
        vertical sep = 2cm,
        horizontal sep = 2.5cm,
        ylabels at=edge left,
      },
        xlabel=$k$,
        ylabel=$ \lambda$,
        yticklabels={},
        xticklabels={},
        height=0.28\textwidth,
        width=0.24\textwidth,
        view = {0}{90},
        colormap/cool,
    ]
      \nextgroupplot[colorbar right, colorbar style = {title=BLEU, width=2.2mm, xshift=-2mm, font=\footnotesize}, yticklabels={}, xticklabels={}, font=\footnotesize]
      \addplot3[matrix plot*] file []  {worst_lstm.dat};
      \nextgroupplot[colorbar right,  colorbar style = {title=BLEU, width=2.2mm, xshift=-2mm, font=\footnotesize}, yticklabels={}, xticklabels={}, font=\footnotesize]
      \addplot3[matrix plot*] file []  {exp_lstm.dat};
      \nextgroupplot[colorbar right,  colorbar style = {title=BLEU, width=2.2mm, xshift=-2mm, font=\footnotesize}, yticklabels={}, xticklabels={}, font=\footnotesize]
      \addplot3[matrix plot*] file []  {best_lstm.dat};
      
       \nextgroupplot[colorbar right, colorbar style = {title=BLEU, width=2.2mm, xshift=-2mm, font=\footnotesize}, yticklabels={}, xticklabels={}, font=\footnotesize]
      \addplot3[matrix plot*] file []  {worst_trans.dat};
      \nextgroupplot[colorbar right,  colorbar style = {title=BLEU, width=2.2mm, xshift=-2mm, font=\footnotesize}, yticklabels={}, xticklabels={}, font=\footnotesize]
      \addplot3[matrix plot*] file []  {exp_trans.dat};
      \nextgroupplot[colorbar right,  colorbar style = {title=BLEU, width=2.2mm, xshift=-2mm, font=\footnotesize}, yticklabels={}, xticklabels={}, font=\footnotesize]
      \addplot3[matrix plot*] file []  {best_trans.dat};
      
  \end{groupplot}
    \node (title) at ($(group c1r1.center)!0.5!(group c1r1.center)+(-3.0cm,2.5cm)$) {\textbf{LSTM}};
    \node (title) at ($(group c1r2.center)!0.5!(group c1r2.center)+(-2.5cm,2.5cm)$) {\textbf{Transformer}};

    \node (title) at ($(group c1r1.center)!0.5!(group c1r1.center)+(0,2.0cm)$) {\textbf{linear}};
    \node (title) at ($(group c2r1.center)!0.5!(group c2r1.center)+(-0.1cm,2.0cm)$) {\textbf{exponential}};
    \node (title) at ($(group c3r1.center)!0.5!(group c3r1.center)+(0,2.0cm)$) {\textbf{sigmoid}};
    
    \node (title) at ($(group c1r2.center)!0.5!(group c1r2.center)+(0,2.0cm)$) {\textbf{linear}};
    \node (title) at ($(group c2r2.center)!0.5!(group c2r2.center)+(-0.1cm,2.0cm)$) {\textbf{exponential}};
    \node (title) at ($(group c3r2.center)!0.5!(group c3r2.center)+(0,2.0cm)$) {\textbf{sigmoid}};    

    \end{tikzpicture}
\caption{Hyperparameter analysis using heatmaps. Better performance is achieved with different hyperparameter pairs over standard scheduled sampling~(bottom row where $\lambda = 0$), thus showing the robustness of our proposed variant. We set $\lambda$ to $\{0, 0.0005, 0.005, 0.1, 0.5\}$, $k$ for the linear schedule to $\{-0.000003, -0.00003, -0.0003, 0.00003, 0.0003\}$, $k$ for the exponential schedule to $\{0.999997, 0.99999, 0.9999, 0.999, 1\}$, and $k$ for the sigmoid schedule to $\{6,000, 8,000, 10,000, 12,000, 15,000\}$.}\label{fig:heatmap}
\end{center}
\end{figure*}
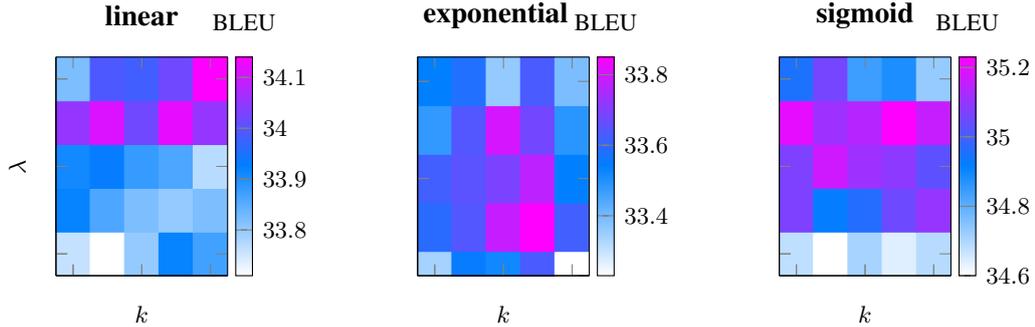

\begin{figure}[!t]
  \begin{center}
    \begin{tikzpicture}
    \begin{groupplot}[group style={
            group size=2 by 1,
            ylabels at=edge left,
            vertical sep= 2cm
        }, 
        height=0.28\textwidth,
        width=0.27\textwidth,
        xmin=0, xmax=80,
        ymin=15, ymax=40,
        ymajorgrids=true,
        xlabel={Sequence Length},
        ylabel style={align=center},
        ylabel={BLEU},
        ylabel shift=-0.5ex,
        xlabel shift=-0.5ex,
        grid style=dashed,
        label style={font=\small},
        title style={yshift=-1ex},
        tick label style={font=\small},
        legend to name=named,
        legend columns=-1
    ]
    \nextgroupplot[align=center, title={LSTM}, style={font=\small}]
     \addplot[color=blue, mark=square] coordinates {(10,27.6)(30,22.8)(50,20.5)(70,19.2)};
    \addplot[color=red, mark=triangle]  coordinates {(10,29.1)(30,25.9)(50,23.8)(70,22.9)};
    \addplot[color=green, mark=triangle]  coordinates { (10,29.9)(30,26.9)(50,26.4)(70,24.9)};
    \nextgroupplot[align=center, title={Transformer}, style={font=\small}]
     \addplot[color=blue, mark=square] coordinates {(10,35.1)(30,33.4)(50,29.8)(70,28.2)};
     \addplot[color=red, mark=triangle]  coordinates {(10,35.4)(30,34.6)(50,31.9)(70,30.9)};
     \addplot[color=green, mark=triangle]  coordinates { (10,35.6)(30,35.2)(50,34.1)(70,32.8)};
    \addlegendentry[color=black]{MLE}
    \addlegendentry[color=black]{SS}
    \addlegendentry[color=black]{SS + EWC}
    \end{groupplot}
    \end{tikzpicture}
    \ref{named}
   \caption{BLEU scores versus output sequence length~(bins of width 20).}\label{fig:ss_ewc_length_ablation}
  \end{center}
\end{figure}
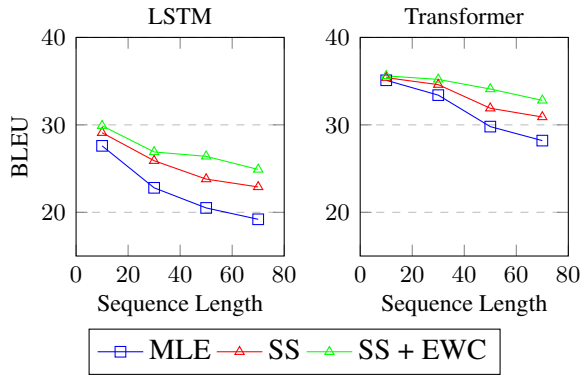

\paragraph{Performance at Different Output Lengths}
Figure~\ref{fig:ss_ewc_length_ablation} shows BLEU scores on the DE-EN test set for different output sequence lengths by grouping them into bins of width 20. We observe that the output quality degrades as the sequence length increases across all training objectives. In particular, MLE-trained models suffer the most from this due to exposure bias. Conversely, our proposed scheduled sampling variant improves the BLEU score across all bins~(compared to MLE and standard scheduled sampling), even for long sequences.

\paragraph{Effect of $\lambda$} Figure~\ref{fig:heatmap} shows the BLEU scores on the DE-EN test set under different values for $k$~(for the annealing schedule) and $\lambda$~(for EWC). We observe that scheduled sampling with EWC outperforms standard scheduled sampling~(bottom row where $\lambda=0$) across all hyperparameter pairs. Thus we conclude that the introduction of $\lambda$ does not increase the complexity of hyperparameter tuning as it improves the robustness of scheduled sampling across annealing schedules and their parameterizations. \citet{Huszar2018ewc} pointed out that EWC is an approximation of the joint optimization of the loss on two datasets, via encapsulating the information from the initial dataset in the model parameters that are fine-tuned. Scheduled sampling (without EWC) also aims at achieving the same goal, i.e.\ optimizing the parameters for generating outputs using both ground-truth prefixes and model-generated ones. In principle, this should be possible to achieve via a well-tuned annealing schedule, however, as Figure~\ref{fig:heatmap} shows, it is substantially easier to tune the hyperparameter controlling the EWC regularizer. Finally, in the theoretical analysis of SEARN which also uses an annealing schedule in a similar manner,~\citet{DBLP:journals/ml/DaumeLM09} argued that the theoretically guaranteed annealing rate would be too slow to be applied in practice.

\section{Related Work}
\paragraph{Imitation Learning for NMT}
Scheduled sampling is an adaptation to recurrent neural networks~(RNNs) and autoregressive models more broadly of~\textsc{DAgger}~\citep{DBLP:journals/jmlr/RossGB11}, a well-known imitation learning~(IL) technique for mitigating exposure bias. Since~\citet{DBLP:conf/nips/BengioVJS15}, a number of research works have adapted standard IL algorithms~\citep{DBLP:journals/ml/DaumeLM09, DBLP:journals/corr/RossB14, DBLP:conf/icml/ChangKADL15} to autoregressive models to address exposure bias.~\textsc{SEARNN}~\citep{DBLP:conf/iclr/LeblondAOL18} is a neural adaptation of~\textsc{SEARN}~\citep{DBLP:journals/ml/DaumeLM09} that computes a local loss for each generated token.~\citet{goyal-etal-2017-differentiable} proposed a differentiable scheduled sampling variant to mitigate the credit assignment problem, while~\citet{xu-etal-2019-differentiable} aimed to fix the time step alignment issue between the oracle and the model-generated tokens. \citet{zhang-etal-2019-bridging} introduced a sequence-level scheduled sampling variant which conditions on model-generated prefixes with the highest BLEU score. A separate family of methods focus on using IL algorithms for non-autoregressive NMT~\citep{wei-etal-2019-imitation, DBLP:conf/nips/GuWZ19, DBLP:conf/icml/WelleckBDC19, DBLP:journals/corr/abs-2011-06868}, and knowledge distillation~\citep{liu-etal-2018-distilling, lin-etal-2020-autoregressive}.      

\paragraph{Non-MLE Training Methods} 
The discrepancy between MLE training and inference has prompted the development of many alternatives training algorithms. Approaches based on reinforcement learning~(RL), such as Actor-Critic~\citep{DBLP:conf/iclr/BahdanauBXGLPCB17} and~\textsc{MIXER}~\citep{DBLP:journals/corr/RanzatoCAZ15} optimize the task-level loss directly.~\textsc{GOLD}~\citep{DBLP:journals/corr/abs-2009-07839} uses off-policy RL to learn from human demonstrations through importance weighting~\citep{hastings1970monte}. Reward augmented maximum-likelihood~(RAML)~\citep{DBLP:conf/nips/NorouziBCJSWS16} maximizes the likelihood of sequences with respect to the exponentiated reward distribution. Minimum Risk Training~(MRT)~\citep{och-2003-minimum, smith-eisner-2006-minimum, shen-etal-2016-minimum} optimizes model parameters by minimizing the expected loss directly with respect to the task-level loss. Other non-MLE training methods include energy-based models~\citep{DBLP:conf/iclr/DengBOSR20}, beam-search optimization~(BSO)~\citep{wiseman-rush-2016-sequence}, and unlikelihood training~\citep{DBLP:conf/iclr/WelleckKRDCW20}.

\paragraph{Mitigating Catastrophic Forgetting}
Despite progress, various state-of-the-art neural architectures in NLP fail to retain previously learned knowledge due to catastrophic forgetting~\citep{DBLP:journals/corr/abs-1901-11373}. To this end, many approaches have been proposed to overcome this limitation in a plethora of NLP tasks, such as NMT~\citep{miceli-barone-etal-2017-regularization, khayrallah-etal-2018-regularized, saunders-etal-2019-domain, thompson-etal-2019-overcoming}, reading comprehension~\citep{DBLP:conf/ijcnn/XuZJL20}, fact verification~\citep{Thorne2020ElasticWC}, and visual question answering~\citep{DBLP:conf/aaai/PerezSVDC18, greco-etal-2019-psycholinguistics}. Another line of research works investigates techniques which address catastrophic forgetting during the fine-tuning stage of deep pretrained language models~\citep{arora-etal-2019-lstm, chronopoulou-etal-2019-embarrassingly, DBLP:conf/nips/ChenWFLW19, chen-etal-2020-recall, DBLP:journals/corr/abs-2009-09139}.

\section{Conclusion}
In this work, we conduct systematic analyses to examine the strengths and weaknesses of scheduled sampling. By applying LRP to quantify the influence of the input sequence and the prefix during prediction, we show that models trained with scheduled sampling increase their reliance on the former to address exposure bias. However, we also demonstrate that as a side-effect, this leads to output degradation due to catastrophic forgetting when the prefix generated by the model is correct. Accordingly, we propose using EWC to better balance mitigating exposure bias with retaining performance. Our EWC-regularized scheduled sampling variant alleviates catastrophic forgetting and significantly improves performance over MLE and scheduled sampling baselines on four IWSLT'14 and WMT'14 translation datasets.

\section*{Limitations}
Even though we show that it is easy to tune the hyperparameter controlling the EWC regularizer across different annealing schedules and their parameterization (Figure~\ref{fig:heatmap}), in terms of runtime, we empirically observe that estimating the Fisher matrix diagonal for EWC, incurs additional training costs on top of running scheduled sampling. Furthermore, we did not assess whether the improvements we report are in any way dependent on properties of the language pairs we considered in our experiments. Thus, further testing is needed to establish whether our approach would transfer to other translation datasets.

\section*{Acknowledgements}
The authors wish to thank Ferenc Husz\'{a}r and Bill Byrne for their helpful comments and feedback. Michalis Korakakis is supported by the Cambridge Commonwealth, European and International Trust and the ESRC Doctoral Training Partnership for Social Sciences.

\bibliography{anthology,custom}
\bibliographystyle{acl_natbib}

\end{document}